\documentclass[runningheads]{llncs}
\usepackage{graphicx}
\usepackage{amsfonts}
\begin{document}
\title{Exploring Bayesian Deep Learning for Urgent Instructor Intervention Need in MOOC Forums %\thanks{Supported by organisation x.}
}
\author{Jialin Yu \and
Laila Alrajhi \and
Anoushka Harit \and
Zhongtian Sun \and
Alexandra I. Cristea \and
Lei Shi}
% \author{Information Removed for Double Blinded Review}

\institute{Department of Computer Science, Durham University, Durham, UK \\
\email{\{jialin.yu,laila.m.alrajhi,anoushka.harit,
%\}@durham.ac.uk}
%\email{\{
zhongtian.sun,alexandra.i.cristea,lei.shi\}@durham.ac.uk}}

% \institute{Information Removed for Double Blinded Review}

%
\maketitle              
% typeset the header of the contribution
\begin{abstract}
% current 187 words 
% https://wordcounter.net/

Massive Open Online Courses (MOOCs) have become a popular choice for e-learning thanks to their great flexibility. However, due to large numbers of learners and their diverse backgrounds, it is taxing to offer real-time support. Learners may post their feelings of confusion and struggle in the respective MOOC forums, but with the large volume of posts and high workloads for MOOC instructors, it is unlikely that the instructors can identify all learners requiring intervention. This problem has been studied as a Natural Language Processing (NLP) problem recently, and is known to be challenging, due to the imbalance of the data and the complex nature of the task. In this paper, we explore for the first time Bayesian deep learning on learner-based text posts with two methods: Monte Carlo Dropout and Variational Inference, as a new solution to assessing the need of instructor interventions for a learner's post. We compare models based on our proposed methods with probabilistic modelling to its baseline non-Bayesian models under similar circumstances, for different cases of applying prediction. The results suggest that Bayesian deep learning offers a critical uncertainty measure that is not supplied by traditional neural networks. This adds more explainability, trust and robustness to AI, which is crucial in education-based applications. Additionally, it can achieve similar or better performance compared to non-probabilistic neural networks, as well as grant lower variance.

\end{abstract}
\section{Introduction}

MOOCs are well-known for their high dropout rates \cite{alamri2019predicting}\cite{alamri2020mooc}. Whilst learners may discuss their problems in the forums before actually dropping out, the sheer volume of posts renders it almost impossible for instructors to address them. Thus, many of these urgent posts are overlooked or discarded. Hence, a few researchers proposed \cite{almatrafi2018needle}\cite{chandrasekaran2015learning} automated machine learning models for need prediction based on learners' posts in MOOC forums. Such an approach would allow instructors to identify learners who require urgent intervention, in order to, ultimately, prevent potential dropouts (see our recent research, where we have shown only 13\% of learners passing urgent intervention messages complete the course \cite{alrajhi2021urgency}). 

More recently, techniques for applying deep neural networks to interpret texts from the educational field have emerged \cite{hernandez2019systematic}, including identifying learners' needs based on their posts in forums \cite{alrajhi2020multidimensional}\cite{guo2019attention}\cite{sun2019identification}. Despite their success, standard deep learning models have limited capability to incorporate uncertainty. One other challenge is that post data is notoriously imbalanced, with urgent posts representing a very low percentage of the overall body of posts - the proverbial 'needle in the haystack'. This tends to make a neural network overfit and ignore the urgent posts, resulting in large variance in model predictions.

To address the above two challenges, we apply Bayesian probabilistic modelling to standard neural networks. Recent advances in Bayesian deep learning offer a new theory-grounded methodology to apply probabilistic modelling using neural networks. This important approach is yet to be introduced in the Learning Analytics (LA) field. 

Thus, the main contributions of this work are:

\begin{enumerate}
    \item We present the first research on how Bayesian deep learning can be applied to text-based LA. Here, the aim is to predict instructor intervention need in the educational domain. 
    \item Hence, we explore, for the first time, not only one, but two Bayesian deep learning methods, on the task of classifying learners' posts based on their urgency, namely Monte Carlo Dropout and Variational Inference.
    \item We show empirically the benefits of Bayesian deep learning for this task and we discuss the differences in our two Bayesian approaches.
    \item We achieve competitive results in the task and obtain a lower variance when training with small size data samples.
    \item We apply this approach to text-based processing on posts in MOOCs - a source generally available across all MOOC providers. Thus our approach is widely applicable - generalisable to foresee instructor's intervention need in MOOCs in general and to support the elusive problem of MOOC dropout. 
\end{enumerate}

\section{Related Work}

\subsection{Urgent Intervention Need in MOOCs}

Detection of the need for urgent instructor intervention is arguably one of the most important challenges in MOOC environments. The problem was first proposed and tackled \cite{chaturvedi2014predicting} as a binary prediction task on instructor's intervention histories based on statistical machine learning. A follow-up study \cite{chandrasekaran2015learning} proposed the use of $L1$ regularisation techniques during the training and used an additional feature about the type of forum (thread), besides the linguistic features of posts. Another study \cite{almatrafi2018needle} tried to build a generalised model, using different shallow ML models with linguistic features extracted by NLP tools, metadata and term frequency. In general, this problem was attempted based on two types of data format: text-only \cite{bakharia2016towards}\cite{clavie2019edubert}\cite{guo2019attention}\cite{sun2019identification}\cite{wei2017convolution} or a mixture of text and post features \cite{almatrafi2018needle}\cite{chandrasekaran2015learning}. From a machine learning perspective, both traditional machine learning methods \cite{almatrafi2018needle}\cite{bakharia2016towards}\cite{chandrasekaran2015learning} and deep learning based methods \cite{clavie2019edubert}\cite{guo2019attention}\cite{sun2019identification}\cite{wei2017convolution} were proposed and explored; with more recent studies being in favour of deep neural network-based approaches \cite{hernandez2019systematic}. However, one critical problem for deep neural networks is that they do not offer a robust estimation over the prediction values. Also, we can not perform efficient learning on small sample size data. Thus, in this paper, we explore the benefits of Bayesian deep learning to predict learners who require urgent interventions from an instructor. We use text only features in our study, as it is the first study to explore the benefits of this new approach, and we leave future optimisation for further work.

To the best of our knowledge, this is the first study of Bayesian deep learning methods for learners' urgent intervention need classification. Our research sheds light on a new direction for other researchers in the fields of Educational Data Mining (EDM) and Learning Analytics (LA).

\subsection{Bayesian Neural Networks}

Modern neural networks (NNs) are self-adaptive models with a learnable parameter set $W$. In a supervised learning setting, given data $D={(x_{i}, y_{i})}^{N}_{i=1}$, we aim to learn a function through the neural network $y = f_{NN}(x)$ that maps the inputs $x$ to $y$. A point estimation of the model parameter set $W^{{\ast}}$ is obtained through a gradient based optimisation technique and with a respective cost function. 

Bayesian neural networks (BNNs) \cite{mackay1992practical}\cite{mackay1995probable}\cite{neal2012bayesian}, alternatively, consider the probability of the distribution over the parameter set $W$ and introduce a prior over the neural network parameter set $P(W)$. The posterior probability distribution $P(W \mid D)$ is learnt in a data-driven fashion through Bayesian inference. This grants us a distribution over the parameter set $W$ other than a static point estimation, which allows us to model uncertainty in the neural network prediction. In the prediction phase, we sample model parameters from the posterior distribution i.e. $w \sim P(W \mid D)$ and predict results with $f_{NN}^{w}(x)$ for the corresponding $y$. We marginalise the $w$ samples and obtain an expected prediction. Due to the complexity and non-linearity of neural networks, an exact inference for BNNs is rarely possible, hence various approximation inference methods have been developed \cite{gal2016dropout}\cite{graves2011practical}\cite{hernandez2015probabilistic}\cite{hernandez2015predictive}. The most widely adopted approximation method is the Monte Carlo Dropout \cite{gal2016dropout}, with applications in natural language processing, data analytics and computer vision  \cite{gal2016theoretically}\cite{kendall2015bayesian}\cite{kendall2017uncertainties}\cite{zhu2017deep}\cite{xiao2019quantifying}. In this paper, we adopt the same idea and use Monte Carlo Dropout \cite{gal2016dropout} to approximate the neural network as a BNN.

\subsection{Variational Inference}
\label{related_work: vi}

Variational inference (VI) \cite{bishop2006pattern}\cite{jordan1999introduction}\cite{wainwright2008graphical} is a general framework for Bayesian statistical modelling and inference under a maximum likelihood learning scheme. It introduces an unobserved random variable as the generative component to model the probabilistic uncertainty. Given fully observed data $D={(x_{i}, y_{i})}^{N}_{i=1}$,  we consider them as random variables and use capital letters $X$ and $Y$ to represent them. The unobserved random variable introduced with VI is denoted as $Z$ and passes the information from $X$ to $Y$. It can be marginalised out with Bayes' rule as:

\begin{equation}
  P(X, Y; \theta) = \sum_{Z} P(X, Y, Z; \theta) = P(Y \mid Z; \theta)P(Z \mid X; \theta)
\end{equation}
Under a mean-field assumption \cite{tanaka1999theory} over the unobserved random variable $Z$, we can factorise it as:

\begin{equation}
  P(Z; \theta) = P(z_{1}, ..., z_{N}; \theta) =  \prod_{i=1}^{N} P(z_{i}; \theta)
\end{equation}
Hence for each pair of data $x$ and $y$, the maximum likelihood learning delivers the following objective with respect to $\theta$:

\begin{equation}
  \log P(y|x;\theta)=\log \int_{z} P(y|z;\theta)P(z|x; \theta)dz 
\end{equation}
Given observed data $D={(x_{i}, y_{i})}^{N}_{i=1}$, we can not directly model the distribution of unobserved $z$ and hence the probability distribution $P(y|z;\theta)$ is intractable for data-driven models, such as neural networks. With VI, an additional variational parameter $\phi$ with its associated variational family distribution $q(z; \phi)$ is introduced, to approximate the real probability $P(y|z;\theta)$. During the learning process, we minimise the distance between $q(z; \phi)$ and $P(y|z;\theta)$ through the Kullback–Leibler divergence, a term that measures the distance between two probability distributions. Hence, the learning of the intractable probability distribution problem is converted to an optimisation problem over the evidence lower bound (ELBO), where $\mathbb{D}_{KL}$ refers to the Kullback–Leibler divergence:

\begin{equation}
    \log P(y|x;\theta) \geq \mathcal{L}(ELBO) =  \mathbb{E}_{q(z; \phi)}[\log P(y|z; \theta)]-\mathbb{D}_{KL}[q(z; \phi)||p(z|x; \theta)]
\end{equation}
 VI was initially developed to solve a specific class of modelling problems where conditional conjugacy is presumed, and variational parameter $\phi$ is updated through closed-form coordinate ascent \cite{ghahramani2000propagation}. However, conditional conjugacy is not practical in most of the real world problems; thus further advancements \cite{blei2017variational}\cite{hoffman2013stochastic}\cite{kingma2013auto}\cite{ranganath2014black}\cite{zhang2018advances} extend VI to large scale datasets and non-conjugate models.

\section{Methodology}

In this section, we first introduce the baseline model built based on recurrent neural networks (RNNs) and an attention mechanism. Then we present our two approaches for applying Bayesian deep learning with our baseline model: 1) Monte Carlo Dropout and 2) Variational Inference.

\subsection{Baseline Deep Learning Model}
\label{methodology: base}

\begin{figure}[h]
\vskip -0.2in
\centering
\includegraphics[width=0.50\linewidth]{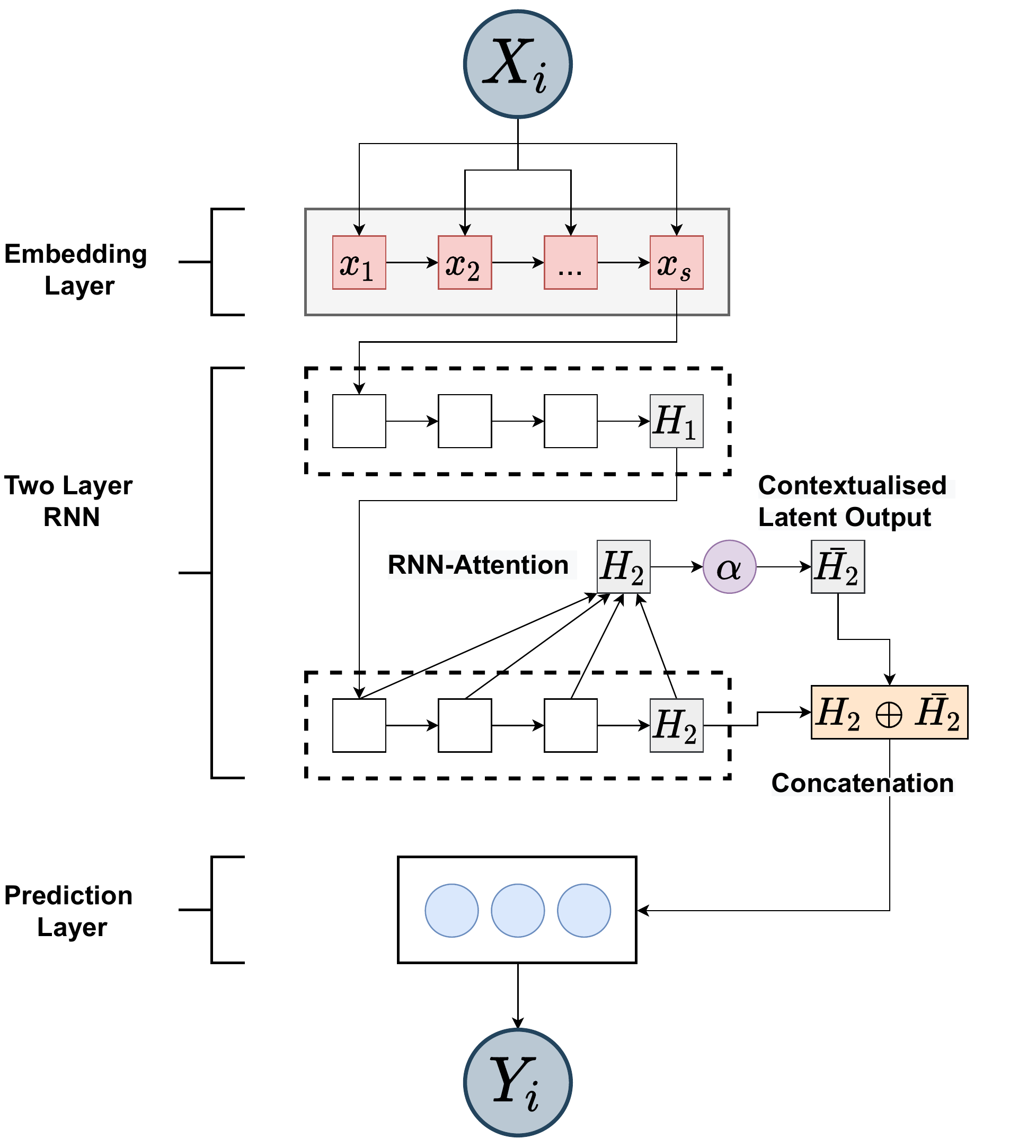}
\vskip -0.1in
\caption{Model architecture for baseline model (operation $\oplus$ refers to the concatenation).}
\label{fig: basemodel}
\vskip -0.3in
\end{figure}

In this section, we first introduce our non-Bayesian model, which serves as our baseline model. The model consists of three different components: an embedding layer, a two-layer recurrent neural network (RNN), and a prediction layer. We use attention based on the output of the RNNs to create a contextual representation over the RNN hidden outputs and then concatenate it with the last layer RNN outputs. The model architecture is presented in Figure \ref{fig: basemodel}.
\newline
Given the data $D={(x_{i}, y_{i})}^{N}_{i=1}$, where  each sentence $x_{i}$ consists of a sequence of tokens $x_{i}^{1}, x_{i}^{2}, ..., x_{i}^{s}$ where $s$ denotes the sequence length. For our baseline model, given a sentence $x_{i}$, we first pass it through the embedding layer and obtain a sequence of word embeddings:

\begin{equation}
    E = (emb(x_{i}^{1}), emb(x_{i}^{2}), ..., emb(x_{i}^{s}))
\end{equation}
where $emb$ is the embedding function we used for our experiment with $d$ dimensions. Here, $x_{i}^{m}$ denotes the $m^{th}$ word in the sentence $x_{i}$. For the initial sentence $x_{i} \in \mathbb{R}^{s \times 1}$, we derive a sentence $x_{i} \in \mathbb{R}^{s \times d}$ after the embedding layer. Then we feed this as a sequence input through a two-layer long-short-term memory (LSTM) model as in \cite{sun2019identification}. The initial hidden state $h_{0}$ is set to $0$ and we calculate the sequence of hidden states as:
\begin{equation}
    h_{m} = LSTM(h_{m-1}, x_{i}^{m})
\end{equation}
where we have $m=1,...,s$. The last layer of hidden states provides a sequence output $H \in \mathbb{R}^{s \times h}$, where $h$ here represents the hidden dimension size. In order to utilise the contextual information through the LSTM encoding process, we calculate the attention score $\alpha$ based on the last hidden state outputs $H_{2}$ ($H_{2}=h_{s}$) and each hidden state in the sequence of $H$, as:
\begin{equation}
    \alpha_{m} = \frac{H_{2}*h_{m}}{\sum_{m=1}^{s}H_{2}*h_{m}}
\end{equation}
Then we calculate the contextual $\bar{H_{2}}$ as:
\begin{equation}
    \bar{H_{2}} = \sum_{m=1}^{s} \alpha_{m}h_{m}
\end{equation}
Finally, we concatenate them and feed them through a fully connected layer with the output dimension equal to the number of classes for our task. This fully connected layer is represented as a prediction layer in Figure \ref{fig: basemodel}.

\subsection{Model Uncertainty with Monte Carlo Dropout}
% The Bayesian neural network aims to find the posterior distribution of $P(W \mid D)$ instead of a static value in the given the data $D={(x_{i}, y_{i})}^{N}_{i=1}$.

In this section, we present how to convert our baseline model into a Bayesian neural network. With Monte Carlo Dropout \cite{gal2016dropout}, we only need to use the dropout technique \cite{srivastava2014dropout} before each layers containing the parameter set $W$. In our case, we add a dropout layer after the first and second LSTM layers, as well as after the fully connected layer, which takes the input as the concatenation of $\bar{h_{2}}$ and $h_{2}$. 
\newline
\begin{figure}[h]
\vskip -0.2in
\centering
\includegraphics[width=\linewidth]{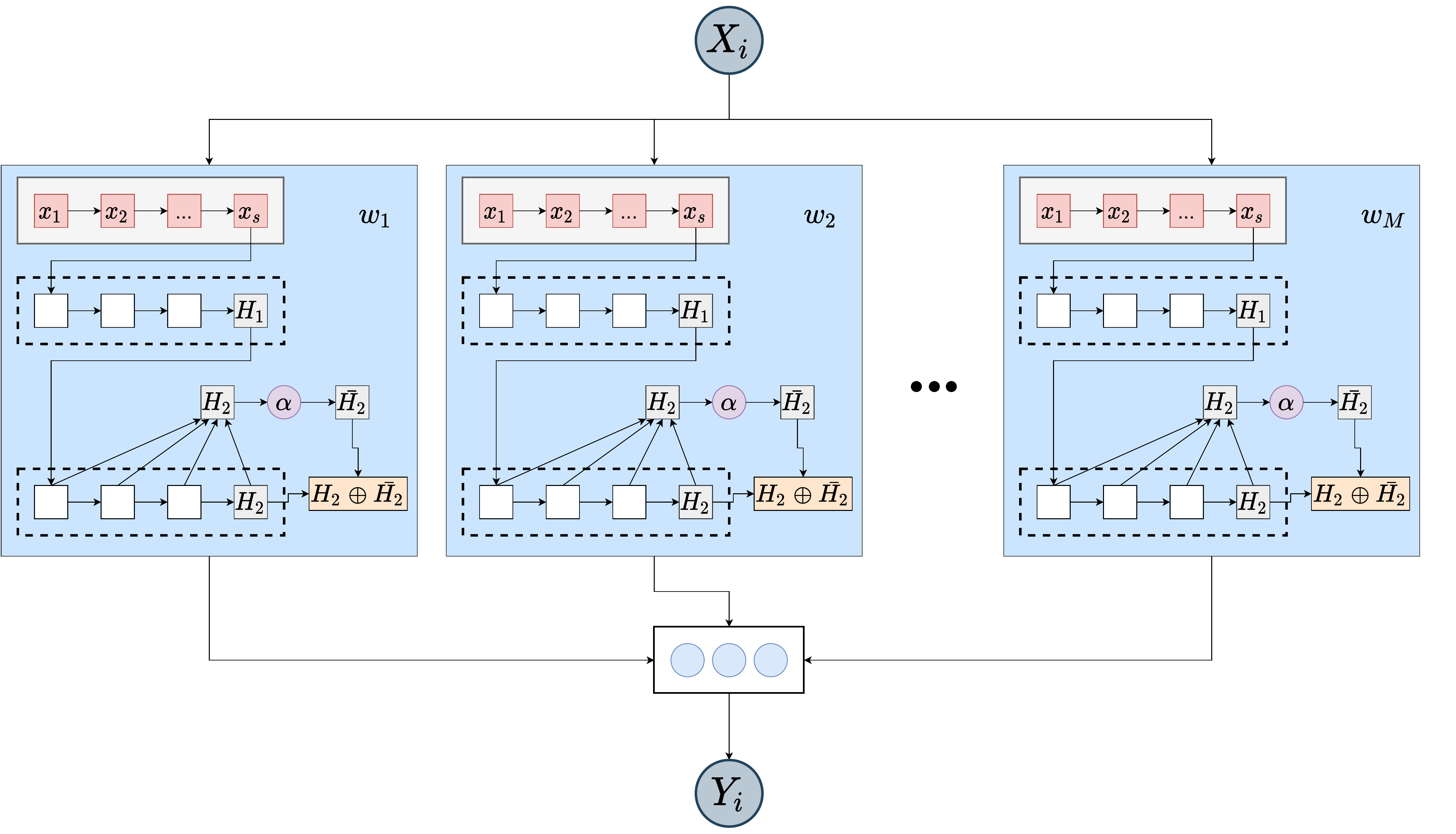}
\vskip -0.1in
\caption{A demonstration of the Monte Carlo Dropout in the test phase. We run the model for $M$ times for $M$ different prediction results and then calculate their average as the prediction layer output.}
\label{fig: BNN}
\vskip -0.3in
\end{figure}
Compared with the standard dropout technique, which works as a regularisation technique in the training phase only, the Monte Carlo Dropout technique requires the dropout layer to be activated in both training and testing phases. This allows the standard neural network model to work as a BNN \cite{gal2016dropout}. Each dropout  works as a sample of $w$ from its probabilistic distribution space and hence allows us to measure the uncertainty of the model, as shown in Figure \ref{fig: BNN}. In the testing phase, we predict the output through sampling $M$ times \cite{gal2016dropout} and the expectation of $y$ can be calculated as:
\begin{equation}
    \mathbb{E}(y \mid x) \approx \frac{1}{M}\sum^{M}_{i=1} f_{NN}^{w_{i}}(x)
\end{equation}
We use this expectation as the final logits value and in our experiment we use a total sample $M$ of 50 as in \cite{kendall2017uncertainties}.

\subsection{Model Uncertainty with Variational Inference}

\begin{figure}[ht]
\vskip -0.3in
\centering
\includegraphics[width=0.5\linewidth]{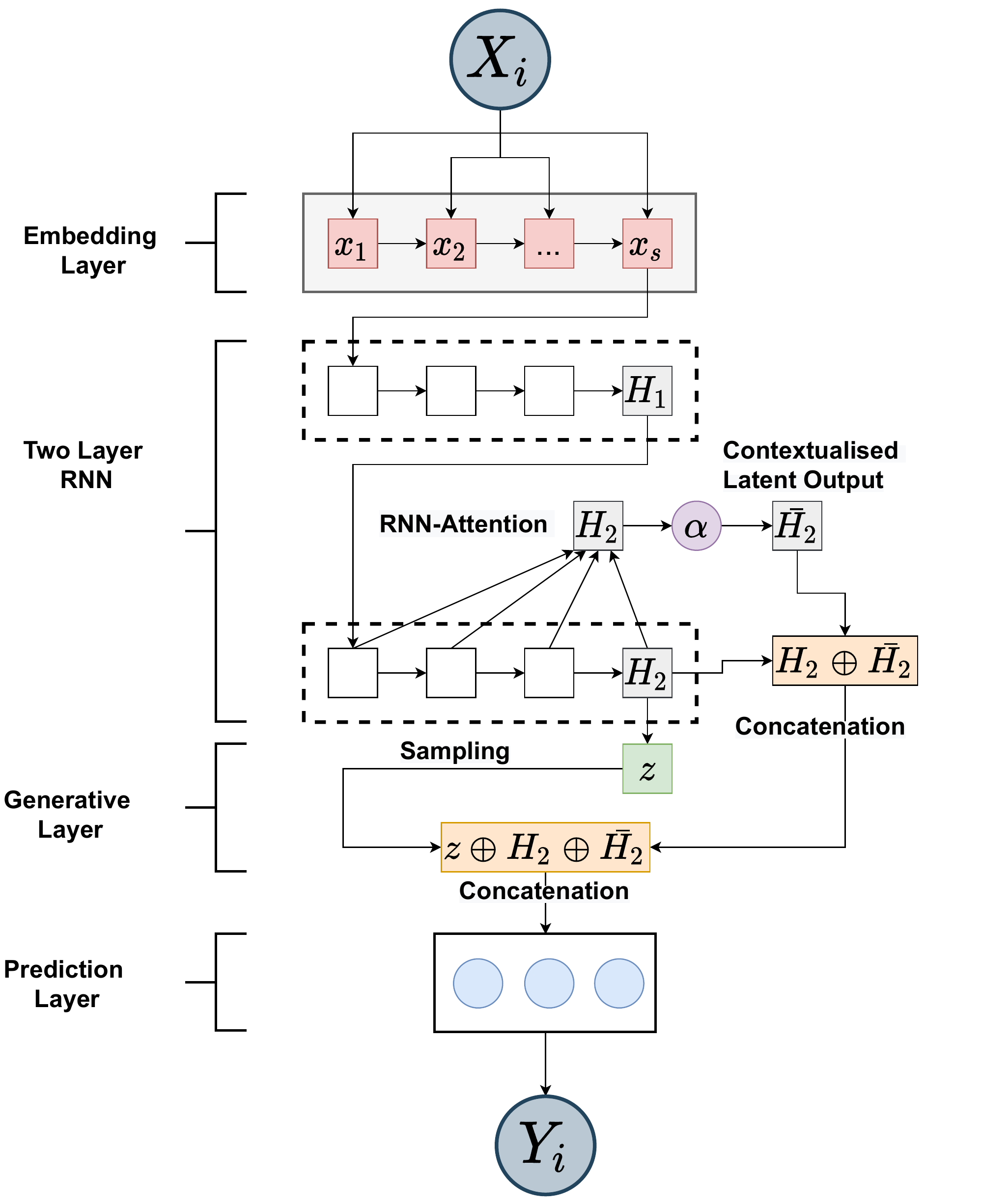}
\vskip -0.1in
\caption{Model architecture for the VI model.}
\label{fig: vi_model}
\vskip -0.5in
\end{figure}

As discussed in the section \ref{related_work: vi}, Variational Inference (VI) introduces an additional random variable $z$ with probability distribution $q(z; \phi)$ to the original model. This variational family $q(z; \phi)$ here approximates the posterior distribution $P(z \mid x; \theta)$ as $q(z|x, y; \phi)$. The model architecture is presented in Figure \ref{fig: vi_model}. Following \cite{miao2016neural}, we define $q_{\phi}(z|x, y)$ as:

\begin{equation}
    q(z|x, y; \phi) = \mathcal{N}(z|\mu_{\phi}(x, y), \textup{diag}(\sigma^{2}_{\phi}(x, y)))
\end{equation}
we have:
\begin{equation}
    \mu_{\phi}(x, y)=l_{1}(\pi_{\phi})
\end{equation}
and:
\begin{equation}
    \log\sigma_{\phi}(x, y)=l_{2}(\pi_{\phi})
\end{equation}
where:
\begin{equation}
    \pi_{\phi} = g_{\phi}(H_{2}, f_{y}(y))
\end{equation}
where $f_{y}(y)$ is an affine transformation from output $y \in \mathbb{R}^{1}$ to a vector space size $s_{y} \in \mathbb{R}^{z}$. The $H_{2}$ is the final latent state output of the second LSTM network layer as stated in section \ref{methodology: base}. The latent variable $z \in \mathbb{R}^{h}$ can be reparameterised as $z=\mu+\sigma\cdotp\epsilon$, known as the "reparameterisation trick" \cite{kingma2015variational} with sample $\epsilon \sim \mathcal{N}(0, \mathbb{I})$. For the conditional distribution $p_{\theta}(z|x)$, we can model it as:

\begin{equation}
    p(z|x; \theta) = \mathcal{N}(z|\mu_{\theta}(x), \textup{diag}(\sigma^{2}_{\theta}(x)))
\end{equation}
where we have:
\begin{equation}
    \mu_{\theta}(x)=l_{3}(\pi_{\theta})
\end{equation}
\begin{equation}
    \log\sigma_{\theta}(x)=l_{4}(\pi_{\theta})
\end{equation}
and: 
\begin{equation}
    \pi_{\theta} = g_{\theta}(H_{2})
\end{equation}
where $l_{1}, l_{2}, l_{3}$ and $l_{4}$ are four affine transformation functions. Since both $p(z|x; \theta)$ and $ q(z|x, y; \phi)$ are multivariate Gaussian distributions, this allows us to have a closed-form solution for the Kullback–Leibler (KL) divergence term  \cite{kingma2013auto}. For the reconstruction term $\log p(y \mid z; \theta)$ with Monte Carlo approximation \cite{miao2016neural}, the final reconstruction loss can be calculated as:

\begin{equation}
    \mathbb{E}_{q(z)}[\log p(y|z; \theta)] \approx \frac{1}{M}\sum _{m=1}^{M}\log p(y|z_{m} \oplus H_{2} \oplus \bar{H_{2}}; \theta)
\end{equation}
where $\oplus$ denotes the concatenation operation and M is the number of samples from the posterior distribution $z$. We use a single sample of $M=1$ during training based on  \cite{kim2018tutorial} and $M=20$ during testing based on \cite{miao2016neural}. In the training phase, $z$ is sampled from $q(z|x, y; \phi)$ and in the test phase, from $p(z|x; \theta)$.

\section{Experiments}
\subsection{Dataset}

Here, we used the benchmark posts dataset from the Stanford MOOC forum \cite{agrawal2015youedu}, containing 29604 anonymised posts collected from 11 different courses. Each post is manually labelled by three independent human experts and with agreements for the gold label. Apart from the text content, each post is evaluated based on six categories, amongst which urgency, which is the one we used here. Its range is 1 to 7; with 1 meaning no reason to read the post and 7 meaning extremely urgent for instructor interventions. An example urgent message is "I hope any course staff member  can help us to solve this confusion asap!!!"; whilst a non-urgent would be "Good luck to everyone.". See more details on their website\footnote{https://datastage.stanford.edu/StanfordMoocPosts/}. Similar to \cite{guo2019attention}, we convert the problem of detecting urgent posts to a binary classification task. A threshold of 4 is used as in \cite{guo2019attention} to create two need categories as: 1) \textit{Need for urgent intervention (value $>$ 4) with label 1}; and 2) \textit{No need for intervention (values $\leq$ 4) with label 0}. This allows us to obtain a total of 29,597 posts, with 23,991 labelled as 0 and 5,606 labelled as 1. We tokenise the text and create a vocabulary based on a frequency-based cutoff \cite{luong2015effective} and use the special token $<pad>$ for padding and the unknown token $<unk>$ for out-of-vocabulary words. We initialise the embedding layer with a 300-dimensional GloVe vector \cite{pennington2014glove} if found in the pre-trained token list. 

\subsection{Experiment Setup and Evaluation}

In this paper we have implemented 3 different models: a baseline model (\textit{Base}), as shown in Figure \ref{fig: basemodel}; a baseline model converted to a Bayesian neural network through Monte Carlo Dropout (\textit{MCD}), as shown in Figure \ref{fig: BNN}; and a baseline model with variational inference (\textit{VI}), as shown in Figure \ref{fig: vi_model}. For the evaluation, we report mean accuracy; F1 score, Precision score, Recall score for all three models under each class (the higher the better); and entropy based on the prediction layer \cite{kendall2017uncertainties} \cite{xiao2019quantifying} (the lower the better).

We conduct two sets of experiments. For the first set, we follow the setup in \cite{alrajhi2020multidimensional}. At each run of the experiment, we randomly split this data into training and testing sets each with a ratio of 80\% and 20\%, respectively, with stratified sampling on a random state. In the second set of experiments, we use less training examples, since the intervention case is rare compared with non-intervention, and we compare the robustness of our model given smaller size samples and we use a split of 40\%, 60\% for training and testing. The results for the two experiments are reported in Table \ref{tab:result1} and Table \ref{tab:result2}, respectively, and we run both experiments 10 times. In Table \ref{tab:result1}, we report the best run of the model and in Table \ref{tab:result2}, we report the mean and variance. All the evaluation metrics results reported here in this paper are based on test dataset only. In the first table, we use bold text to denote the results that outperform results in \cite{alrajhi2020multidimensional} and in the second table, we we use bold  to denote results outperforming the (\textit{Base}) model.

\section{Results and Discussions}

The results are presented in Table \ref{tab:result1}. The baseline model (Base) performs competitively against a strong model \cite{alrajhi2020multidimensional}, especially in the recall and F1 score for the 'urgent' class and the precision score for the 'non-urgent' class. For the Monte Carlo Dropout (MCD) and Variational Inference (VI) models, we achieve better performance in these measurements against the baseline model (Base). Importantly, as an indication of the uncertainty measurement, we note that the entropy dropped for the MCD model. In Table \ref{tab:result2}, we can see that Bayesian deep learning methods generally achieve similar or better performance compared to the non-Bayesian base model, but hold lower variance and lower entropy against small sample size data. This is often the case in real life scenarios, where the label 'need intervention' is scarce. A probabilistic approach works as a natural regularisation technique, when neural network models are generally over-parameterised. We can conclude that Bayesian deep learning mitigates this issue of over-parametrisation with lower variance and entropy. This is especially clear for the VI methods. The result from a Wilcoxon test shows that, compared with the Base model, the experiment results of the VI model are statistically significant at the $\textit{.05}$ level, with $\textit{p=.022}$ for the entropy value and with $\textit{p=.007}$ for the recall, in the 'urgent' case. Comparing MCD and VI models, the latter achieves better performance in most metrics, as shown in both tables, especially with a higher recall score. The recall score is preferable to precision in this task, where we have a comparatively small number of positive examples. However, the implementation of MCD models is more accessible to researchers interested in introducing uncertainty into their neural networks. This should be considered in using them in practice.

\begin{table}[t]
\small
\caption{Results compare baseline model and Bayesian deep learning approach in accuracy, precision, recall and F1 score. }
\label{tab: results}
\vskip -0.20in
\begin{center}
\begin{tabular}{p{4em}|p{5em}p{4em}|p{5em}p{4em}p{3em}|p{5em}p{4em}p{3em}}
% \hline
%  & &  & \textbf{Dataset} &  & \\
\hline
\hline
 \multicolumn{3}{c}{} & \multicolumn{3}{c}{\textbf{Non-urgent (0)}} & \multicolumn{3}{c}{\textbf{Urgent (1)}} \\
\hline
\hline
& \textbf{Accuracy} & \textbf{Entropy} & \textbf{Precision} & \textbf{Recall} & \textbf{F1} & \textbf{Precision} & \textbf{Recall} & \textbf{F1}\\
\hline
\textbf{Text \cite{alrajhi2020multidimensional}} & .878 & - & .90 & .95 & .93 & .73 & .56 & .64 \\
\hline
\textbf{Base} & \textbf{.883} & .095 & \textbf{.937} & .918 & .927 & .677 & \textbf{.738} & \textbf{.697} \\
\hline
\hline
\textbf{MCD} & .883 & \textbf{.085} & \textbf{.939} & .915 & .926 & .675 & \textbf{.742} & \textbf{.698} \\
\hline
\textbf{VI} & .873 & .103 & \textbf{.940} & .901 & .919 & .644 & \textbf{.752} & .687 \\
\hline
\hline
% \textbf{Model} \\
\end{tabular}
\end{center}
\label{tab:result1}
\vskip -0.2in
\end{table}

\begin{table}[t]
\small
\caption{Results compare mean and variance of Deep learning and Bayesian deep learning approach based on 10 runs.}
\label{tab:results2}
\vskip -0.20in
\begin{center}
\begin{tabular}{p{3em}|p{5em}p{4em}|p{5em}p{4em}p{3em}|p{5em}p{4em}p{3em}}
% \hline
%  & &  & \textbf{Dataset} &  & \\
\hline
\hline
 \multicolumn{3}{c}{} & \multicolumn{3}{c}{\textbf{Non-urgent (0)}} & \multicolumn{3}{c}{\textbf{Urgent (1)}} \\
\hline
\hline
& \textbf{Accuracy} & \textbf{Entropy} & \textbf{Precision} & \textbf{Recall} & \textbf{F1} & \textbf{Precision} & \textbf{Recall} & \textbf{F1}\\
\hline
\textbf{Base} & .870+-.0039 & .1126+-.0041 & .930+-.0039 & .908+-.0088 & .918+-.0030 & .645+-.0159 & .707+-.0215 & .664+-.0052 \\
\hline
\textbf{MCD} & .869+-\textbf{.0013} & \textbf{\underline{0.101}}+-.0326 & .929+-.0128 & .908+-.0319 & .917+-.0104 & .652+-\textbf{.0574} & .703+-.0693 & .660+-\textbf{.0042} \\
\hline
\textbf{VI} &  .867+-\textbf{.0019} & \textbf{\underline{0.078}}+-.0296 & .924+-\textbf{.0028} & \textbf{\underline{.910}}+-\textbf{.0058} & .916+-\textbf{.0017} & \textbf{.642}+-\textbf{.0093} & .680+-\textbf{.0164} & .649+-\textbf{.0034} \\
\hline
\hline
% \textbf{Model} \\
\end{tabular}
\end{center}
\small
\label{tab:result2}
\vskip -0.4in
\end{table}

\section{Conclusion}

Identifying the need of learner interventions for instructors is an extremely important issue in MOOC environments. In this paper, we have explored the benefits of a Bayesian deep learning approach to this problem for the first time. We have implemented two different approaches to Bayesian deep learning, namely Monte Carlo dropout and variational inference. Both offer a critical probabilistic measurement in neural networks. We have demonstrated the effectiveness of both approaches in decreasing the epistemic uncertainty of the original neural network and granting equivalent or even better performance. We have thus provided guidelines for researchers interested in building safer, more statistically sound neural network-based models in the Learning Analytics (LA) field. Entropy measures a classifiers' confidence level. In intelligent tutoring systems, high confidence (thus low entropy) is essential. With Bayesian deep learning, we turn NN models into probabilistic models, allowing more explanability and trust. For future research, these can be extended and applied in more areas.

% for an empty page
\clearpage
% \newpage

\bibliographystyle{splncs04}
\bibliography{its}
\end{document}